\documentclass[10pt,twocolumn,letterpaper]{article}
\usepackage{revision}
\usepackage{cvpr}
\makeatletter
\@namedef{ver@everyshi.sty}{}
\makeatother
\usepackage{times}
\usepackage{epsfig}
\usepackage{graphicx}
\usepackage{amsmath}
\usepackage{amssymb}
\usepackage{subfigure}
\usepackage{rotating}
\usepackage{multirow}
\usepackage{float}
\usepackage{bm}
\usepackage{tikz}

\makeatletter
\renewcommand{\@thesubfigure}{\hskip\subfiglabelskip}
\makeatother

\usepackage{indentfirst}
\usepackage{booktabs}
\makeatletter
\renewcommand{\paragraph}{%
  \@startsection{paragraph}{4}%
  {\z@}{1.8ex \@plus 1ex \@minus .2ex}{-1em}%
  {\normalfont\normalsize\bfseries}%
}
\makeatother

\usepackage[pagebackref=true,breaklinks=true,letterpaper=true,colorlinks,bookmarks=false]{hyperref}

\cvprfinalcopy 
\def\cvprPaperID{515} 

\begin{document}

\title{Learning to Calibrate Straight Lines for Fisheye Image Rectification}

\author{Zhucun Xue, Nan Xue, Gui-Song Xia\thanks{Corresponding author: guisong.xia@whu.edu.cn}, Weiming Shen\\
{\em CAPTAIN-LIESMARS, Wuhan University, Wuhan, 430079, China}\\
{\tt\small \{zhucun.xue, xuenan, guisong.xia, shenwm\}@whu.edu.cn}
}

\maketitle
\vspace{-3mm}
\begin{abstract}
This paper presents a new deep-learning based method to simultaneously calibrate the intrinsic parameters of fisheye lens and rectify the distorted images. Assuming that the distorted lines generated by fisheye projection should be straight after rectification, we propose a novel deep neural network to impose explicit geometry constraints onto processes of the fisheye lens calibration and the distorted  image  rectification.
In addition, considering the nonlinearity of distortion distribution in fisheye images, the proposed network fully exploits multi-scale perception to equalize the rectification effects on the whole image.
To train and evaluate the proposed model, we also create a new large-scale dataset labeled with corresponding distortion parameters and  well-annotated distorted lines. Compared with the state-of-the-art methods, our model achieves the best published rectification quality and the most accurate estimation of distortion parameters on a large set of synthetic and real fisheye images.


\end{abstract}

\vspace{-2mm}
\section{Introduction}
Fisheye cameras have been widely used in many computer vision tasks~\cite{szeliski1997creating,xiong1997creating,bertozzi2000vision,huang20176,alhaija2018augmented} because of their large field of view (FOV), however, the images taken by fisheye cameras always suffer from severe geometric distortion simultaneously.
When processing the geometric toward vision systems equipped with fisheye lens,
calibrating the intrinsic parameters is usually the first step we should do to rectify the distorted images.

\subsection{Motivation and Objective}
Early work viewed the calibration of fisheye cameras as an optimization problem by fitting the relationship between 2D/3D calibration patterns from images with different viewpoints~\cite{kannala2006generic,Grossberg2001A,Sturm2004A,ScaramuzzaMS06}. But these methods usually demand pre-prepared calibration patterns and extra manual operations, and even often involve heavy off-line estimations, which seriously limit their usage scenarios in real applications.
\begin{figure}[t]
\includegraphics[width=0.24\linewidth]{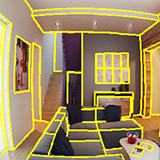}
\includegraphics[width=0.24\linewidth]{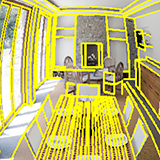}
\includegraphics[width=0.24\linewidth]{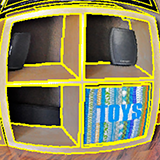}
\includegraphics[width=0.24\linewidth]{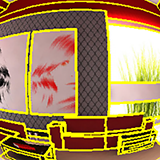}
\vspace{.5mm}
\includegraphics[width=0.24\linewidth]{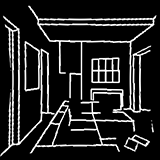}
\includegraphics[width=0.24\linewidth]{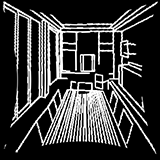}
\includegraphics[width=0.24\linewidth]{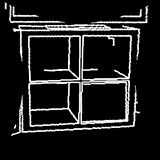}
\includegraphics[width=0.24\linewidth]{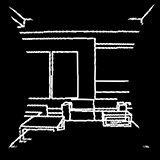}
\vspace{.5mm}
\includegraphics[width=0.24\linewidth]{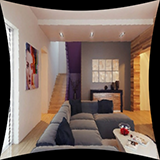}
\includegraphics[width=0.24\linewidth]{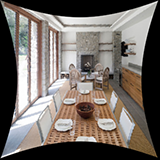}
\includegraphics[width=0.24\linewidth]{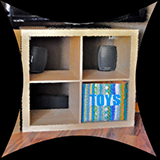}
\includegraphics[width=0.24\linewidth]{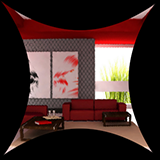}
\caption{Learning to calibrate straight lines for fisheye image rectification. {\bf \bm{$1^{st}$}-Row:} original fisheye images overlaid by detected {\em distorted lines} which should be straight after calibration; {\bf \bm{$2^{nd}$}-Row:} rectified straight lines; {\bf \bm{$3^{rd}$}-Row:} rectified image. }
\vspace{-15pt}
\label{fig:1}
\end{figure}
To overcome these limitations and toward an automatic self-calibration solution, subsequent investigations were proposed to detect geometric objects ({\em e.g., conics and lines}) from a single image and further exploit their correspondences in 3D world ~\cite{devernay2001straight,zhang2015line,barreto2009automatic,melo2013unsupervised,bukhari2013automatic,aleman2014automatic}. These approaches have reported promising performances on fisheye camera calibrations only when specified geometric objects in fisheye images can be accurately detected. Nevertheless, it is worth noticing that the involved detection of geometric objects in fisheye images itself is another challenging problem in computer vision.
Recently, alternative approaches have been proposed based on deep convolutional neural networks (CNNs)~\cite{rong2016radial,yin2018fisheyerecnet}.
Avoiding the difficult to directly detect the geometric objects, these methods tried to learn more discriminative visual features with CNNs to rectify the distorted image. Although aforementioned methods have reported the state-of-the-art performances on fisheye image rectifications, as well as avert the difficulties of detecting geometric objects,
the geometry characteristics in  fisheye calibration task are not fully exploited using CNNs.

Regardless of the difficulties to detect geometric objects in fisheye images, one should observe that the explicit scene geometries are still the strong constrains to rectify distorted images. It is of great interest to investigate how to apply the fundamental geometric property under the pinhole camera model, \ie, {\em the projection of straight line from space to the camera should be a line}~\cite{devernay2001straight}, to fisheye image calibration networks. As shown in Fig.~{\ref{fig:1}}, we propose a novel networks which further exploit this explicit scene geometries, aiming at solving the challenges of fisheye camera calibration and image rectification by a deep neural network simultaneously.

\begin{figure*}[t]
\centering
\includegraphics[width=0.78\linewidth]{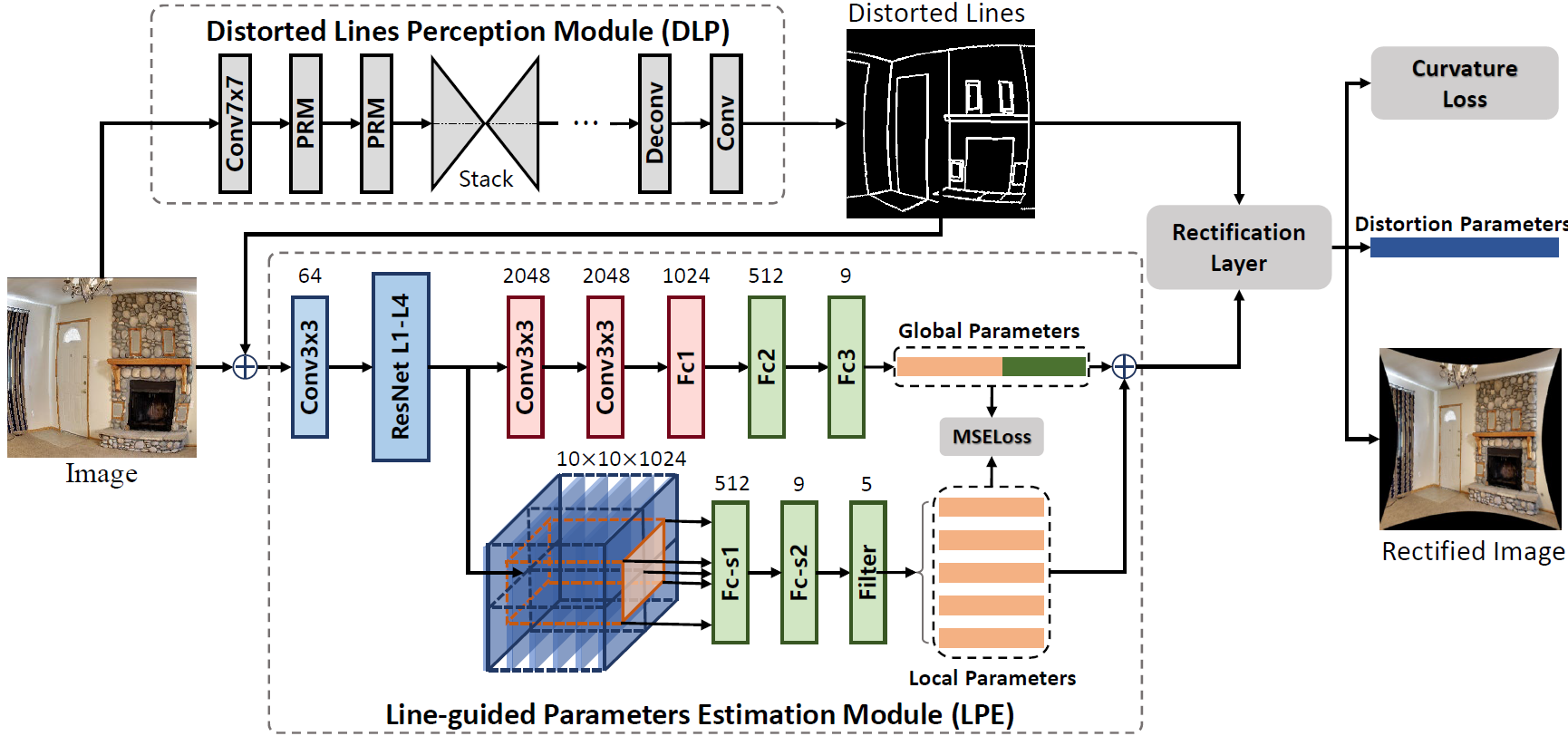}
\caption{Architecture of the overall system. The whole network architecture consists of three parts: line-guided parameter estimation module (LPE), {\em distorted line} segments perception module (DLP), and a rectification layer. DLP could detect the map of curves which should be straight line in rectified image, and take the ouput from DLP and RGB fisheye image into the LPE to estimate the global and local fisheye distortion parameters. The distortion parameters are used in rectification layer to implement the curvature constraint.
}
\label{fig:net}
\vspace{-5pt}
\end{figure*}

\vspace{-1mm}
\subsection{Overview of Our Method}
The above-mentioned challenges motivate following two issues: (1) how to design a powful CNNs to explicitly depict the scene geometry in fisheye images. (2) how to train a deep network effectively and efficiently using geometric information.
To address these problems, we propose to first train a neural network to detect {\em distorted lines} that should be straight after being rectified in fisheye images , and then feed these detected {\em distorted lines} into another deep sub-network to recover the distortion parameters of fisheye camera.
As shown in Fig.~\ref{fig:net}, our network includes three main modules as follows:
\vspace{-1mm}
\begin{itemize}
    \item {\bf Module for detecting distorted straight lines.} This module is designed to extract {\em distorted lines} in the given input fisheye image, which are supposed to be straight in an expected rectified image. Some examples of detected {\em distorted lines} are displayed in the first row of Fig.~\ref{fig:1}. 
\vspace{-1mm}
    \item {\bf Module for recovering the distortion parameters.} Fed with detected {\em distorted lines} and the original fisheye image, this module attempts to recover the distortion parameters of the fisheye lens. In particular, multi-scale perceptron is desined to eliminate the nonlinearity of distortion distribution in fish-eye images by combining both the local and global learning.
\vspace{-1mm}
    \item {\bf Rectification module.}
    This differentiable rectification module serves as a connector between distortion parameters and geometric constraint.
    As shown the second row in Fig.~\ref{fig:1}, the line map without distortion which are rectified from the {\em distorted lines} in the first row are displayed.
\end{itemize}
\vspace{-1mm}
These three modules are trained by minimizing the loss function composed of three terms: a multi-scale constraint of global and local perception on distortion parameters, as well as a curvature constraint on detected {\em distorted lines}.
Since all of the calibration and rectification steps are modeled with one deep neural network, it is naturally to train it in an end-to-end manner.
To get a better performance of proposed network, well-annotations of {\em distorted lines} and distortion parameters are required for every fisheye image during the training phase.
Thus, we create a new synthetic dataset for the fisheye lens calibration by converting the wireframe dataset~\cite{huang2018learning} to \emph{distorted Wireframe collections} (D-Wireframe) and the 3D model repository~\cite{song2016ssc} to \emph{fisheye SUNCG collections} (Fish-SUNCG).
In detail, the D-wireframe collections is created by distorting the perspective images with randomly generated distortion parameters, and the Fish-SUNCG collections is built by rendering the formation of real fisheye lens in 3D virtual scenarios.

\vspace{-1mm}
\subsection{Related Work}
In the past decades, many researches have been devoting themselves to fisheye calibration and distortion correction studies. Earlier works attempted to estimate the distortion parameters by correlating the detected 2D/3D features in specific calibration field~\cite{kannala2006generic,Grossberg2001A,Sturm2004A,ScaramuzzaMS06,aliaga2001accurate,barreto2005geometric,toepfer2007unifying}. However, it is costly to build such calibration fields of large-scale and high-precision as well as it usually turns to be laborious and time-consuming to manually label every calibration pattern.
By contrast, self-calibration methods which rely on structural information detected in distorted images~\cite{devernay2001straight,zhang2015line,barreto2005geometric,melo2013unsupervised,bukhari2013automatic,aleman2014automatic,Peter2006Self} require less manual work and are more efficient.
Devernay~\etal~\cite{devernay2001straight} proposed that the straight line segments in the real world should maintain their line property even after the projection of fisheye lens. Along this axis, Bukhari~\etal~\cite{bukhari2013automatic} recently proposed to use an extended Hough Transform of lines to correct radial distortions. With similar assumption, the `plumb-lines' have used to rectify fisheye distortions~\cite{zhang2015line,melo2013unsupervised,aleman2014automatic} .
However, their correction effects are limited by the accuracy of geometric object detecting results.
Our work in this paper also follows the same assumption as suggested in~\cite{devernay2001straight}, while we propose a deep convolutional neural network to handle the aforementioned problems and generate a more accurate result of {\em distorted lines} extraction in fisheye images.

To mitigate the difficulty of detecting geometric objects in distorted images, the deep learning methods were proposed~\cite{rong2016radial,yin2018fisheyerecnet} which imposed the representational features learned by CNNs to the processes of fisheye calibration and image rectification.
Among them, the FishEyeRecNet~\cite{yin2018fisheyerecnet} proposed an end-to-end CNNs which introduce scene parsing semantic into the rectification of fisheye images. It has reported promising results, but it is still not clear which kind of high-level geometric information learned from their networks are important for fisheye image rectification.
Moreover, The works~\cite{zhang2015line,melo2013unsupervised,aleman2014automatic} explicit geometry like `plumb-lines' are very efficient for distortion corrections, but how to encode them with CNNs in an effective way is still an open problem.

Another topic closely related to our work is {\em distorted lines} extraction in fisheye images. Various arc detection methods and optimizing strategies have been utilized in the calibrating process~\cite{brauer2001new,bazin2007rectangle,bukhari2013automatic,aleman2014automatic,zhang2015line}, but they are not robust to detect arcs especially in the environments with noises or texture absence. Although recent deep learning based methods~\cite{xie2015holistically,liu2017richer,Bertasius2014DeepEdge,Hwang2015Pixel, Piotr2015Fast} show a promising performance on edge detection, none of them is well qualified to deal with {\em ditorted lines} in fisheye images.

\vspace{-1mm}
\subsection{Our Contributions}
In this paper, we propose a novel end-to-end network to calibrate fisheye lens and rectify  distorted images simultaneously with further exploit of geometric constraints. Specifically, we make the following three contributions:
\begin{itemize}
\vspace{-1.5mm}
\item We proposed an end-to-end CNNs to \emph{impose explicit geometry constraints} onto the process of fisheye lens calibration and distorted image rectification, which achieves the state-of-the-art performance.
\vspace{-1.5mm}
\item
Multi-scale perception is designed to balance the nonlinear distribution of distortion effects in fish-eye images. And more robust distortion parameters obtained from global and local learning, so as to achieve better rectification effect.
\vspace{-1.5mm}
\item We construct a new large-scale fisheye dataset to train networks and to evaluate the effectiveness and efficiency of fisheye image rectification methods.
\end{itemize}

\vspace{-2mm}
\section{General Fisheye Camera Model}
Given a normal pinhole camera with focal length $f$ , the perspective projection model can be written as $r = f\tan\theta$,
where $r$ indicates the projection distance between the principal point and the points in the image, and $\theta$ is the angle between the incident ray and the camera's optical axis.
While, fisheye lens violates this perspective projection model~\cite{miyamoto1964fish,basu1995alternative}, and has been often approximated by a general polynomial projection model~\cite{kannala2006generic}, \ie,
\begin{align}
r(\theta)=\sum\nolimits_{i=1}^n k_i\theta^{2i-1},  \quad n=1, 2, 3, 4, \ldots
\label{eq:2}
\end{align}
Usually, this model can accurately approximate the image formation of fisheye lenses when $n$ reaches $5$~\cite{kannala2006generic}.

Given a 3D scene point $\mathbf P_c := (x_c,y_c,z_c)^T \in \mathbb{R}^3$ in the camera coordinate system, it will be projected into the image plane with $\mathbf{p}_d := (x_d,y_d)^T\in \mathbb{R}^2$ refracted by the fisheye lens and $\mathbf p :=(x,y)^T \in \mathbb{R}^2$ through perspective lens without distortion. The correspondence between $\mathbf p_d$ and $\mathbf p$ can be expressed as, $\mathbf p_d = r(\theta) (\cos{\varphi}, \sin{\varphi})^T$,
with $\varphi=\arctan{((y_d-y)/(x_d-x))}$ indicating the angle between the ray that connects the projected point and the center of image and the $x$-axis of the image coordinate system.
Assuming that the pixel coordinate system is orthogonal, we can get pixel coordinates $(u,v)$ converted by image coordinates $\mathbf p_d$ as
\begin{align}
\dbinom{u}{v}=\begin{pmatrix}
m_u&0\\
0&m_v
\end{pmatrix}
\dbinom{x_d}{y_d}+\dbinom{u_0}{v_0}
\label{eq:project}
\end{align}
The principal point of fisheye image is represented as $(u_0,v_0)$, and $m_u$,$m_v$ describe the number of pixels per unit distance in horizontal and vertical direction respectively.

By using Eq.~\eqref{eq:project}, the distortion  effect of fisheye images can be rectified once we can get the parameters $K_d = (k_1,k_2,k_3,k_4,k_5,m_u,m_v,u_0,v_0) $. Therefore, we are going to accurately estimate the parameters $K_d$ for every given fisheye image and simultaneously eliminate the distortion in this paper.

\section{Deep Calibration and Rectification Model}
In this section, we mainly exploit the relationship between scene geometry of {\em distorted lines} and the corresponding distortion parameters of fisheye images by CNNs, and learn mapping functions from raw input fisheye image to the the rectified image.

\subsection{Network Architecture}
As shown in Fig.~\ref{fig:net}, our network is mainly composed of \emph{distorted lines perception module} (DLP) which solves the problem of {\em distorted lines} extraction, \emph{line-guided parameter estimation module} (LPE) which provides estimated distortion parameters $K_d$, as well as \emph{rectification module} which serves as connector between geometry and distortion parameters.
For a RGB fisheye image $I$ with size of $H\times W$, the {\em distorted lines} map $h\in\mathbb{R}^{H\times W}$ could be acquired from DLP, and then fed the {distorted line} map $h$ and the original fisheye image $I$ together into LPE to learn the global and local parameters through multi-scale perception. And the rectification module could verify the accuracy of learned parameters $K_d$ by analyzing whether the lines in rectified {\em distorted lines} map $\hat{h}'$ have be straight after being rectified with $K_d$.
Thereafter, we are able to learn the distortion parameters and rectified
images without distortion in the end-to-end manner.


Every training data sample for our network contains: (1) a fisheye image $I$, (2) the ground truth distortion parameters $\hat{K}_{gt}$,
(3) the ground truth of {\em distorted lines} map $\hat{h}$ of image $I$,
(4) the ground truth of the corresponding rectified line map $\hat{h}'$ and
(5) the corresponding (rectified) line segments $L = \left\{\bm{x}_i,\bm{x}_i'\right\}_{i=1}^{K}$of image $I$, where the $\bm{x}_i\in\mathbb{R}^2$ and $\bm{x}_i'\in\mathbb{R}^2$ are two endpoints of a line segment.

\vspace{-1mm}
\paragraph{Distorted Line Perception Module.}
Followed by the recent advances of edge and line segment detection~\cite{xie2015holistically, huang2018learning}, we use the Pyramid Residual Modules (RPM) and Stacked Hourglass network~\cite{Newell2016Stacked} to learn the {\em distorted line} segments map $h \in \mathbb{R}^{H\times W}$ from the input images. In details, we firstly use two RPMs to extract feature maps with size of $\frac{H}{4}\times\frac{W}{4}\times 256$ from input image with size of $H\times W \times 3$. Then, we pass the feature map into $5$ stacked hourglass modules. The resulted features are then upscaled by using two deconvolution layers to get the feature with size of $H\times W\times 16$. In the end, we use a convolutional layer with $1\times 1$ kernel size to predict the {\em distorted lines} map $h$. Excepting for the prediction layer, the Batch-Normalization and ReLU are used for each (de)convolution layer.
The target of line segment map $\hat{h}$ is defined pixelwised by
\begin{equation}\label{eq:line-map}
    \hat{h}(\bm{p}) = \left\{
    \begin{array}{ll}
    d(\bm{l}) & \text{if}~~\bm{p}~\text{is (nearly) on }  \bm{l} \in L,\\
    0 & \text{otherwise},
    \end{array}
    \right.
\end{equation}
where the function $d(\bm{l}_i)$ is read as
$d(\bm{l}_i) = \left\|\bm{x}_i - \bm{x}_i'\right\|_2$.
The resulted map $h$ not only can  indicate if a pixel $\bm{p}$ is passed through a line segment, the predicted length of rectified line segment also implicitly contains the information for the distortion parameters.

\vspace{-1mm}
\paragraph{Line-guided Parameters Estimation Module.}
This module aims at estimating the distortion parameters from images.
As mentioned above, the predicted {\em distorted lines} map could characterize the distortion of fisheye images in some extent. Based on this,
regarding the {\em distorted lines} map $h$ output from DLP module as a geometric guidance providing high-level structural information to LPE, we concatenate it with the input fisheye image together with size of $H\times W \times 4$, as the input of LPE to estimate the mutli-scale distortion parameters.
As shown in Fig.~\ref{fig:net}, we adopt the level 1 to 4 of the ResNet-50~\cite{he2016deep} as backbone of LPE module, and design a bifurcated structure with a global and a local stream respectively to multi-scale perception, considering the nonlinearity of the distortion distribution in the domain of fisheye images.

The global stream treats the entire feature map to estimate the distortion parameters with $2$ convolutional layers and $3$ fully connected (FC) layers. Before the first FC layer of the global stream, we use the global average pooling operator to extract the abstracted global information from images. The last FC layer output a $9$-D vector representing the distortion parameters denoted by $K_g$.

Because of the nonlinearity of the distortion distribution, we explicitly use the cropped feature maps from the output of the backbone to estimate the distortion parameters locally. We divide this sideoutput into five smaller blocks - the central region with size of $6\times6\times1024$ and four $5\times5\times1024$ feature maps of upper left, lower left, upper right and lower right, and then send these five set of sub-feature maps into two FC layers and a linear filter separately to predict the local parameters, denoted by $\{K_{loc}^k\}_{k=1}^5$. The parameter settings of these two FC layers are same as those in global stream, meanwhile the weight of them are shared across these five set of sub-feature maps. Since the parameters $m_u, m_v$ and $u_0, v_0$ are related to the entire image, the linear filter only reserve the first five distortion parameters $k_1,\ldots,k_5$ in the previous output. Each output $K_{loc}^k$ is thus a $5$D vector.

In training phase, the predicted parameters $K_{loc}^k$ is used as a constraint to regularize the prediction of the global stream. And the output of DLP is the averaged distortion coefficients of global and local parameters, denoted as $K_d$.

\vspace{-1mm}
\paragraph{Rectification Module.}
In this module, we take the predicted distortion parameters $K_{d}$ of LPE module as input to rectify the input fisheye image and the corresponding {\em distorted line} segments map from the DLP module.
Suppose that the pixel coordinate in rectified and fisheye images are $\bm{p}=(x,y)$ and $\bm{p}_d=(x_d,y_d)$, their relationship can be read
\begin{equation}\label{eq:rectify}
    \bm{p}_d = \mathcal{T}(\bm{p},K_d)=\begin{pmatrix}
    u_0\\v_0
    \end{pmatrix} + \frac{r(\theta)\bm{p}}{\left\|\bm{p}\right\|_2}
\end{equation}
With the Eq.~\eqref{eq:rectify}, the {\em distorted lines} map and fisheye image can be rectified by using bilinear interpolation.

The significance of the above rectification layer is to explicitly bridge the relation between distortion parameters and geometry structures. The more accurate the estimated distortion parameters, the {\em distorted lines} map will be rectified better.

\subsection{Loss Function and Training Scheme}\label{subsec:train}
In our network, we can end-to-end output the {\em distorted lines} map $h$, estimated distortion parameters $K_{g}$ and $K_{loc}^{k}, k=1,\ldots,5$ as well as the rectified line segment map for every input image $I$. Inspired by the deeply supervised net \cite{DSN2015} and HED  \cite{xie2015holistically}, we make supervision to the outputs of each module.

\vspace{-3pt}
\paragraph{\bf Loss of Distorted Lines Map Learning.}
Considering the fact that {\em distorted line} segments are $0$-measure geometric primitives in 2D images, most of pixels for the target $\hat{h}$ defined in Eq.~\eqref{eq:line-map} will be $0$. In other words, most of pixels will not be passed through any {\em distorted line} segment. For the sake of representation simplicity, the pixels not being on any {\em distorted line} segment are collected to the negative class $\Omega^-$ and the rest of pixels are collected in the positive class $\Omega^+$, with $\Omega^+ = \Omega - \Omega^- $. Then,
we weight these two classes in the loss function as
\begin{align}
\mathcal{L}_{line} = \cfrac{|\Omega^-|}{|\Omega|}\sum_{\bm{p}\in \Omega^+}\mathcal{D}(\bm{p}) +  \cfrac{|\Omega^+|}{|\Omega|} \sum_{\bm{p}\in \Omega^-}\mathcal{D}(\bm{p}),
\label{eq:line_loss}
\end{align}
where $\mathcal{D}(\bm{p})$ is defined as $\mathcal{D}(\bm{p}) = \|h(\bm{p}) - \hat{h}(\bm{p})\|_2^2$.

\vspace{-4pt}
\paragraph{\bf Loss of Distortion Parameter Estimation.}
In the LPE module, we try to learn the distortion parameters with a bifurcate structure, which results the parameters $K_{g}$ and $\{K_{loc}^{k}\}_{k=1}^5$. Ideally, we want the outputs of LPE module close to the ground-truth distortion parameters. For the output $K_{g}$, we define the loss  as
\begin{align}
\mathcal{L}_g = \frac{1}{9}\sum_{i=1}^9 w_i (K_{g}(i) - K_{gt}(i))^2,
\label{eq:global}
\end{align}
where the $K_g(i)$ and $K_{gt}(i)$ are the $i$-th component of predicted parameter $K_g$ and the ground truth $K_{gt}$. The weight $w_i$ is used to rescale the magnitude between different components of distortion parameter.
On the other side of the bifurcate, the loss of parameters estimated by the sub feature maps are defined as
 \vspace{-1pt}
\begin{align}
    \mathcal{L}_{loc}^k = \frac{1}{5} \sum_{i=1}^5 w_i(K_{loc}^k(i) - K_{gt}^k(i))^2,
    \label{eq:local}
\end{align}
where the $K_{loc}^k(i)$ is the $i$-th component of $K_{loc}^k$.

\vspace{-4pt}
\paragraph{\bf Global Curvature Constraint Loss.}
The $\mathcal{L}_g$ and $\mathcal{L}_{loc}$ enforce the network fit the distortion parameters, however, only optimizing them is not enough and prone to get stuck in the local minimums. Meanwhile, the relation between the parameters and the geometry of {\em distorted line} which should be straight in rectified images provides a stronger constraint to boost the optimizing. If the {\em distorted line} is not completely corrected to a straight line, the estimated distortion parameters are not accurate enough, and vice versa. Therefore, we calculate the pixel-errors between the rectified line map by estimated parameters $K_d$ output from LPE and the ground truth of line map as the global curvature constraint loss $\mathcal{L}_c$:
 \vspace{-1pt}
\begin{align}
\mathcal{L}_c = \frac{1}{N}\sum_{\bm{p}_d\in \Omega^+}(\mathcal{F}(\bm{p}_d,K_d)-\mathcal{F}(\bm{p}_d,K_{gt}))^2,
\label{eq:curloss}
\end{align}

\vspace{-3pt}
where $\mathcal{F}$ is the inverse function of $\mathcal{T}$ which described in Eq.~\eqref{eq:rectify}, and $N$ is the number of pixels that belong to the {\em distorted line} segment.

\vspace{-5mm}
\paragraph{Training Scheme.}
The network training procedure consists of two phases.
In the first phase, we train the {\em distorted line} perception module from scratch with the loss function defined in Eq.~\eqref{eq:line_loss}. Once the DLP module is trained, we fix their parameters and then learn the distortion parameters in the second phase. The total loss we used here is defined as
\begin{equation}
\mathcal{L} = \lambda_{g}\mathcal{L}_{g}+\lambda_{loc}\sum_{k=1}^{5}\mathcal{L}_{loc}^k+\lambda_{c}\mathcal{L}_{c},
\label{eq:loss}
\end{equation}
which aims at fitting the parameters and simulating the distortion effect of fisheye during training. The $\lambda_{g}$, $\lambda_{loc}$ and $\lambda_c$ used in Eq.~\eqref{eq:loss} are weight parameters to balance the different terms.


\vspace{-1mm}
\section{Synthetic Dataset for Calibration} \label{section:dataset}
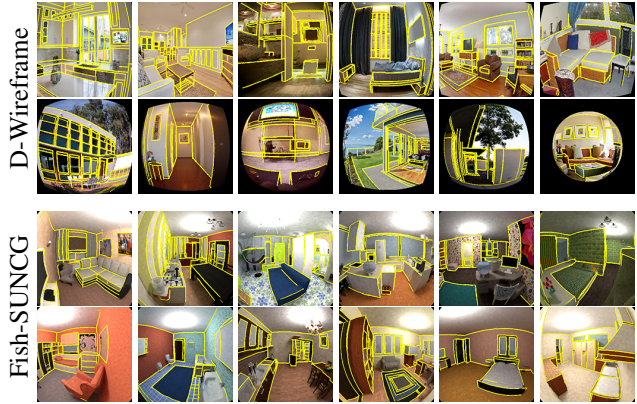
\begin{figure}[t!]
\def\outsize{0.3\linewidth}
\def\xshift{-5mm}
\hspace{\xshift}
\begin{minipage}{0.05\linewidth}
\begin{tikzpicture}
\node[rotate=90] (Ours) at (0,0) {D-Wireframe};
\draw[opacity=0] (-\outsize*0.01,-\outsize*0.5) rectangle (\outsize*0.01,\outsize*0.5);
\end{tikzpicture}
\end{minipage}
\begin{minipage}{1\linewidth}
\includegraphics[width=0.15\linewidth]{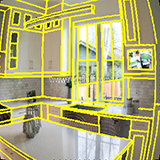}
\includegraphics[width=0.15\linewidth]{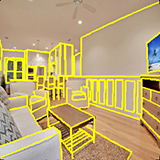}
\includegraphics[width=0.15\linewidth]{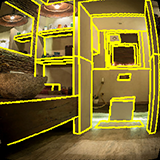}
\includegraphics[width=0.15\linewidth]{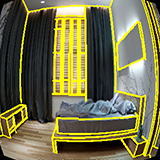}
\includegraphics[width=0.15\linewidth]{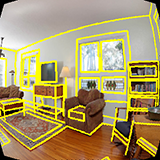}
\includegraphics[width=0.15\linewidth]{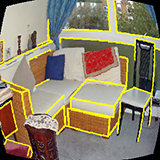}

\includegraphics[width=0.15\linewidth]{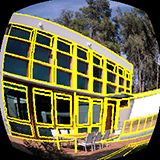}
\includegraphics[width=0.15\linewidth]{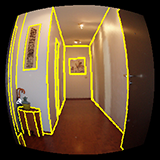}
\includegraphics[width=0.15\linewidth]{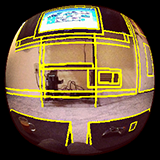}
\includegraphics[width=0.15\linewidth]{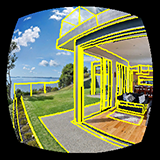}
\includegraphics[width=0.15\linewidth]{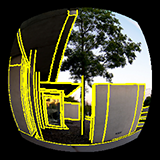}
\includegraphics[width=0.15\linewidth]{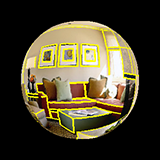}
\end{minipage}

\vspace{2mm}
\hspace{\xshift}
\begin{minipage}{0.05\linewidth}
\begin{tikzpicture}
\node[rotate=90] (Ours) at (0,0) {Fish-SUNCG};
\draw[opacity=0] (-\outsize*0.01,-\outsize*0.5) rectangle (\outsize*0.01,\outsize*0.5);
\end{tikzpicture}
\end{minipage}
\begin{minipage}{1\linewidth}
\includegraphics[width=0.15\linewidth]{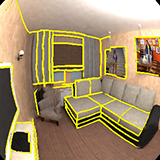}
\includegraphics[width=0.15\linewidth]{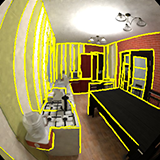}
\includegraphics[width=0.15\linewidth]{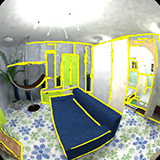}
\includegraphics[width=0.15\linewidth]{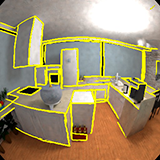}
\includegraphics[width=0.15\linewidth]{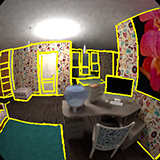}
\includegraphics[width=0.15\linewidth]{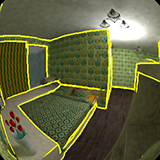}

\includegraphics[width=0.15\linewidth]{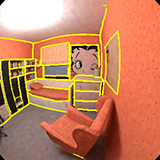}
\includegraphics[width=0.15\linewidth]{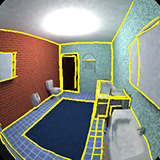}
\includegraphics[width=0.15\linewidth]{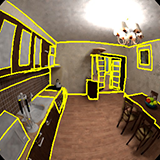}
\includegraphics[width=0.15\linewidth]{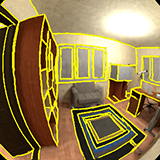}
\includegraphics[width=0.15\linewidth]{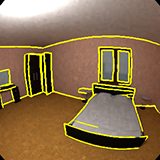}
\includegraphics[width=0.15\linewidth]{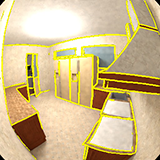}
\end{minipage}
\vspace{1mm}
\caption{Data samples from the \emph{distorted wireframe} (top) and fisheye SUNCG collections (bottom) of our proposed dataset. Every data sample is shown vertically for the original image and corresponding fisheye image.}
\label{fig:dataset}
\vspace{-6mm}
\end{figure}

There still remains a crucial problem for training the proposed neural network which requires real distortion parameters as well as well-annotated distorted and rectified line maps. However, to the best of our knowledge, there is no such large scale dataset that satisfy all above requirements.
Thanks to the recently released wireframe dataset~\cite{huang20176} which has the labeling of straight lines and the large-scale dataset of 3D scenes SUNCG \cite{song2016ssc} which provides diverse semantic 3D scenes, we construct a new dataset with well-annotated 2D/3D line segments $L$ as well as the corresponding distortion parameters $K_{gt}$ for training. The two subsets of our dataset, the distorted wireframe collection (D-Wireframe) from wireframe dataset and the fisheye  SUNCG  collection  (Fish-SUNCG) from 3D model repository, are described below, as shown in Fig.~{\ref{fig:dataset}}.

\begin{figure}[t]
    \centering
    \includegraphics[width=0.32\linewidth]{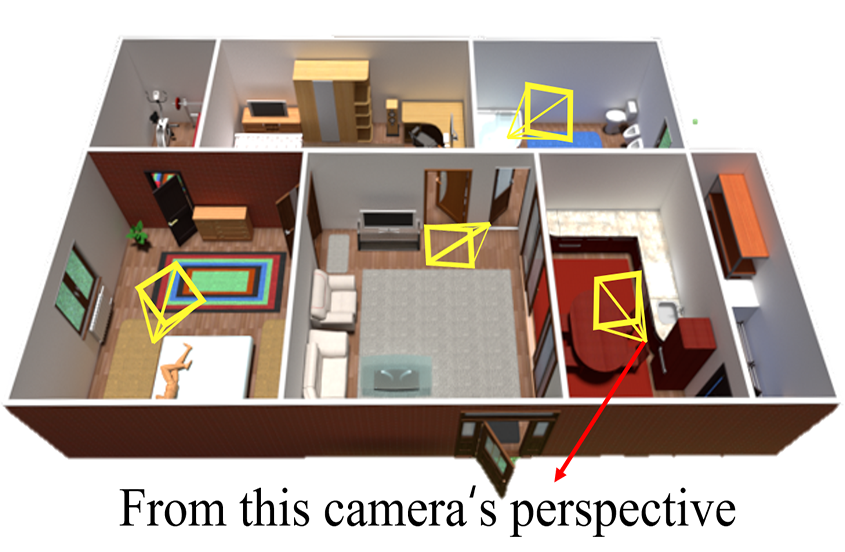}
    \includegraphics[width=0.32\linewidth]{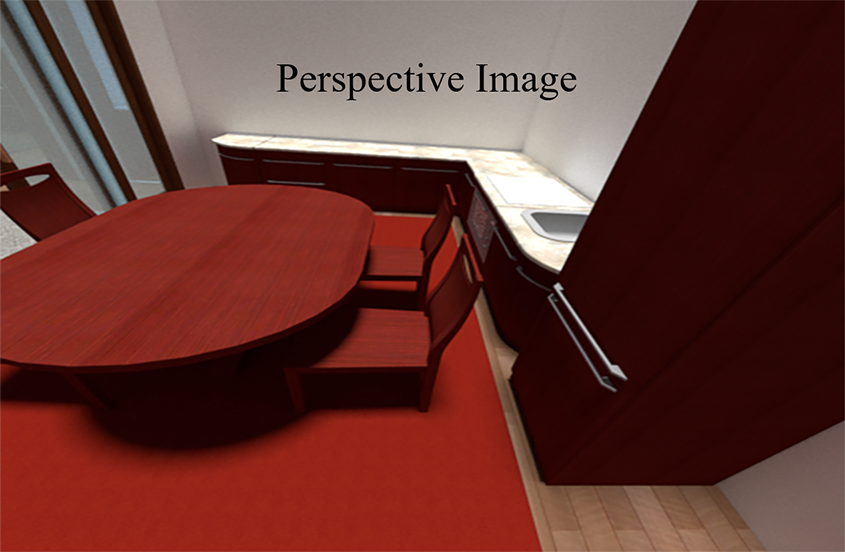}
    \includegraphics[width=0.32\linewidth]{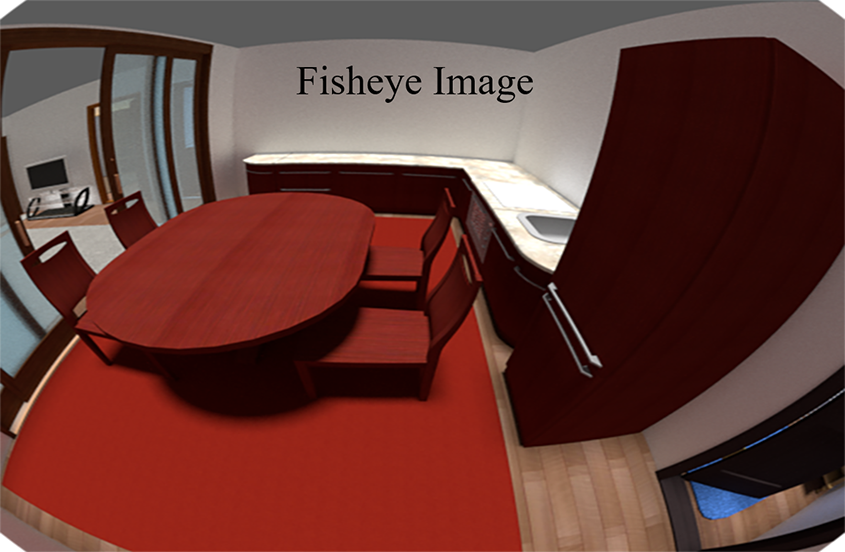}
    \vspace{1mm}
    \caption{Generation schematic diagram of Fish-SUNCG. Equip each camera with perspective lens and fisheye lens respectively. }
    \label{fig:gene_fish}
    \vspace{-5mm}
\end{figure}

\vspace{-1mm}
\paragraph{Distorted Wireframe Collection (D-Wireframe).}
For any image in the wireframes datase proposed by~\cite{huang20176} which contains 5462 normal perspective images marked with straight line segments,
we randomly generate four different sets of distortion parameters $K_i$ to transform this perspective image into fisheye image with different distortion effects by Eq.~\eqref{eq:2}.
Thus, the perspective image and the corresponding line segment annotations can be distorted to the fisheye image with {\em distorted line} segments.
In summary, we generate this collection $\mathcal{D}_{wf}$ and split it into training set and testing set with $20,000$ and $1848$ samples respectively.

\paragraph{Fisheye SUNCG Collection (Fish-SUNCG).}
The D-wireframe collection could provide benefits in terms of diversity and flexibility of fisheye distortion types for the network training. However, artificially distorting the images which taken by perspective cameras cannot fully characterize the fisheye distortion for real scenarios. We address this problem by simulating the image formation for both perspective and fisheye cameras at the same observation positions from the 3D models of SUNCG \cite{song2016ssc} which contains 45K different virtual 3D scenes with manually created realistic room and furniture layouts.
In details, we use the Blender~\cite{blender2014blender} to render images by specifying the camera pose and imaging formation models. The rendering protocol is illustrated in Fig.~{\ref{fig:gene_fish}}.
For the line segments generating, we remove the texture of 3D models to get the wireframe model of 3D objects. After that, we manually remove the edges of wireframe manually to get the line segments that are on the boundary of objects.
Since we are able to control the image formation without metric errors, the data samples can be used to train our network without loss of information. In the end, we generate 6,000 image pairs from 1,000 scenes for training and 300 image pairs from 125 scenes for testing.
This collection is denoted as $\mathcal{D}_{sun}$.

\vspace{-1mm}
\section{Experiments}
\vspace{-1mm}
\subsection{Implementation Details}\label{section:training}
We follow our training scheme described in the Section~\ref{subsec:train}.
We use the distorted fisheye images and the corresponding line segment map of distorted wireframe collect $\mathcal{D}_{wf}$ for training the DLP module in the first step.
After that, we fix the weights of DLP module and train the rest of our network by using the $\mathcal{D}_{wf}$ and $\mathcal{D}_{sun}$ together. The size of input images for our network is set as $320\times 320$ for both training and testing phases.

The weight parameters in Eq.~\eqref{eq:loss} are set to as follow for our experiments: $\lambda_c = 50$, $\lambda_{loc}=\lambda_{g}=1$, and the balance parameters are set to as follow: $W=\{ w_1=0.1,w_2=0.1,w_3=0.5,w_4=1,w_5=1,w_6=0.1,w_7=0.1,w_8=0.1,w_9=0.1 \}$.
The optimization method we used for the training is the stochastic steepest descent method (SGD). The initial learning rate is set to 0.01, and then decrease it by a multiple of 0.1 after 100 epochs. The network will converge after 300 epochs.  And our network is implemented on the PyTorch platform with a single Titan-X GPU device.

\subsection{Evaluation Metrics}
Benefiting from the DLP module of our proposed approach, we are able to compare the effects of eliminating distortion and the performance on recovering line geometry by evaluating the rectified {\em distorted lines} map ${\hat{h}}^{'}$ from rectification module and the ground truth image $\hat{h}$. What's more, the {\em Precision} and {\em Recall} are redefined to quantitatively evaluate the error between ${\hat{h}}^{'}$ and $\hat{h}$. Further, the reprojection error (RPE) is proposed to evaluate the overall rectification effects by measuring the pixels deviation between rectified image and fish image. On the other hand, we also follow the evaluation metrics used in previous works \cite{rong2016radial,yin2018fisheyerecnet} that utilize the peak signal to noise ratio (PSNR) and structure similarity index (SSIM) for evaluating the rectified images.

\vspace{-5pt}
\paragraph{Precision v.s. Recall.}
The precision and recall rate of the line segment map prediction is defined as
\begin{equation}
Precision = |P\cap G|/|P|, \ \ Recall = |P\cap G|/|G|,
\end{equation}
where the $P$ is the set of edge pixels on the rectified line segment map and $G$ is the set of edge pixels in the ground truth of line segment map without distortion.

\vspace{-5pt}
\paragraph{PSNR and SSIM.} These two metrics are widely used to describe the degree of pixel blurring and structure distortion respectively. We use them here for comparing the rectified fisheye images.
In general, the larger the value of PSNR and SSIM, the better the rectification quality.

\vspace{-5pt}
\paragraph{\bf Reprejection Error (RPE).}
This metric is generally used to quantify the distance between an estimate of a 2D/3D point and its true projection.
So we use the real distortion parameters $K_{gt}$ and the estimated ones $K_d$ to rectify the pixels of fisheye image, and get the projection $\mathcal{F}(K_{gt})$ and $\mathcal{F}(K_d)$ respectively, where the $\mathcal{F}$ is the function representation of Eq.~\eqref{eq:curloss}.
The mean square error (MSE) of the RPE in the whole fisheye image is defined by $\frac{1}{N}\sum_{\bm{p}_d\in \Omega}(\mathcal{F}(K_{gt})-\mathcal{F}(k_d))^2$.

\begin{figure}[t]
\begin{rotate}{90}
\quad \ \ \ GT
\end{rotate}
\hspace{0.05cm}
\centering
\includegraphics[width=0.223\linewidth]{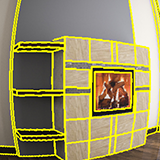}
\includegraphics[width=0.223\linewidth]{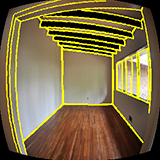}
\includegraphics[width=0.223\linewidth]{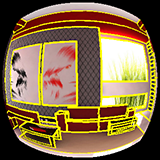}
\includegraphics[width=0.223\linewidth]{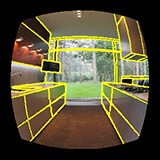}

\begin{rotate}{90}
\ \  Detected
\end{rotate}
\hspace{0.05cm}
\centering
\includegraphics[width=0.223\linewidth]{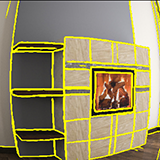}
\includegraphics[width=0.223\linewidth]{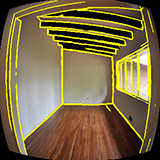}
\includegraphics[width=0.223\linewidth]{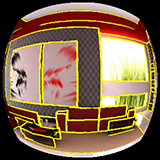}
\includegraphics[width=0.223\linewidth]{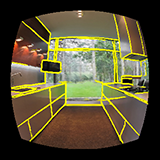}

\caption{{\em Distorted lines} detection results from DLP. {\bf First Row} is ground truth with manual labeling of accurate {\em distorted lines} information. {\bf The Second Row} is the our detection.}
\vspace{-10pt}
\label{fig:line_dection}
\end{figure}

\begin{figure}[t]
\subfigure[][{\scriptsize Distorted Lines}]{
\begin{minipage}[b]{0.19\linewidth}
\includegraphics[width=1\linewidth]{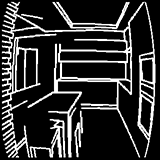}
\includegraphics[width=1\linewidth]{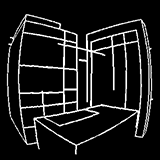}
\end{minipage}
}
\hspace{-1.9ex}
\subfigure[][{\scriptsize Bukhari~\cite{bukhari2013automatic}}]{
\begin{minipage}[b]{0.19\linewidth}
\includegraphics[width=1\linewidth]{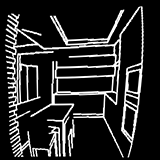}
\includegraphics[width=1\linewidth]{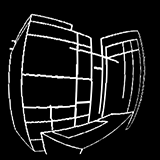}
\end{minipage}
}
\hspace{-1.9ex}
\subfigure[][{\scriptsize AlemnFlores\cite{aleman2014automatic}}]{
\begin{minipage}[b]{0.19\linewidth}
\includegraphics[width=1\linewidth]{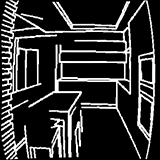}
\includegraphics[width=1\linewidth]{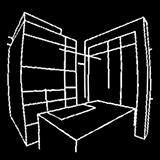}
\end{minipage}
}
\hspace{-1.9ex}
\subfigure[][{\scriptsize Rong~\cite{rong2016radial}}]{
\begin{minipage}[b]{0.19\linewidth}
\includegraphics[width=1\linewidth]{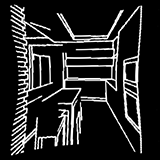}
\includegraphics[width=1\linewidth]{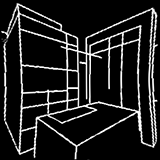}
\end{minipage}
}
\hspace{-1.9ex}
\subfigure[][{\scriptsize Ours}]{
\begin{minipage}[b]{0.19\linewidth}
\includegraphics[width=1\linewidth]{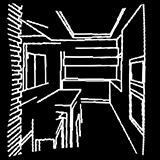}
\includegraphics[width=1\linewidth]{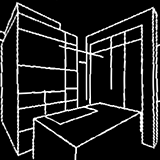}
\end{minipage}
}
\caption{Rectify the distorted lines. First column is the detection results by DLP. Other columns is the rectification results by different
methods.}
\vspace{-1mm}
\label{fig:input_line}
\end{figure}
\begin{figure}[t!]
\vspace{-2mm}
\centering
\includegraphics[width=0.6\linewidth]{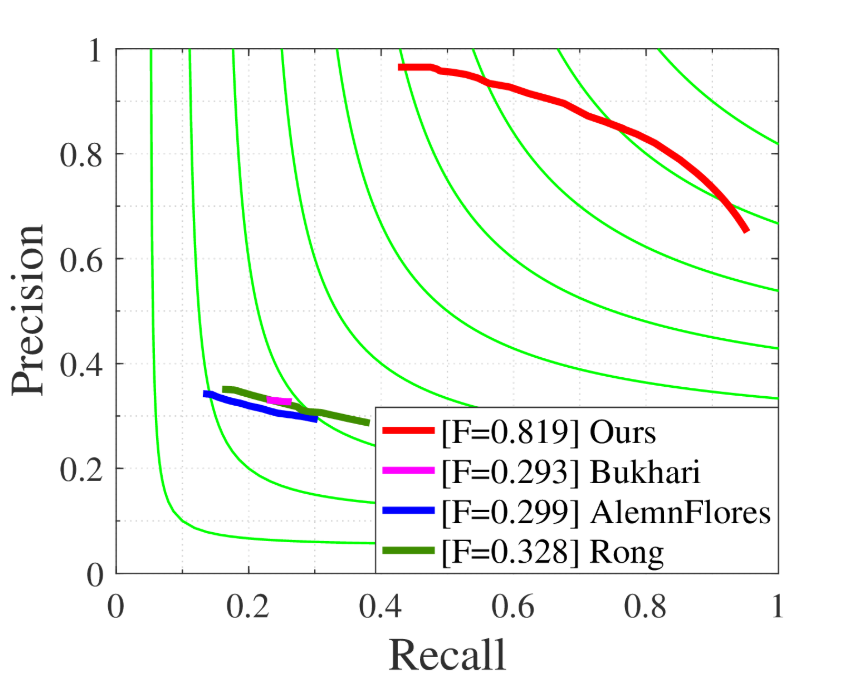}
\caption{The precision-recall curves of different rectification methods for the line map rectification~\cite{bukhari2013automatic,aleman2014automatic,rong2016radial}.}
\vspace{-10pt}
\label{fig:PR}
\end{figure}

\subsection{Comparison with State-of-the-art Methods}
Our network mainly includes {\em distorted lines} detection in DLP and distortion parameters estimation in DPE, aiming at performing better rectification effects.

For DLP, the highly accurate detection of {\em distorted lines} output from this module is the premise of accurate distortion parameters learning. As shown in Fig.~{\ref{fig:line_dection}}, the detection results by our method from DLP show excellent performance as well as close to the ground truth in visual effects.
Instead of directly evaluating the result of {\em distorted line} detection, we jointly evaluate the performance of {\em distorted lines} detection and the accuracy of parameters estimation compared with existing state-of-the-art methods~\cite{bukhari2013automatic,aleman2014automatic,rong2016radial}, and the details will be discussed in the following.

For DPE, the effective use of geometric constraints is the key to guarantee the rectification effects. According to previous analysis, the evaluation for the rectified line segment map can explicitly illustrate the geometric accuracy of rectification. In other words, if the rectified {\em distorted lines} map still exists curved geometry or has deviation from the ground truth, it shows that the learned distortion parameters are not accurate enough. In visual effect, we show the geometric rectification of the {\em distorted lines} map which output from the rectification module intuitively in Fig.~{\ref{fig:input_line}} to verify our network actually has the ability of recovering the straight line characteristics. The results show that the rectified line map through our network is indeed straightened, while those rectified by other methods are still distorted in some extent and it proves the validity of the geometric constraint in our network. Further more, we report the precision and recall curves in Fig.~{\ref{fig:PR}} to show the comparison in quantitative. Obviously, our method is far superior to other methods~\cite{bukhari2013automatic,aleman2014automatic,rong2016radial} in terms of line geometric structure recovery, and achieves the best result (F-value =.819).

\begin{table}[t!]
\centering
\caption{Comparisons with the state-of-the-arts, using the PSNR, SSIM and reprojection error (RPE) calculated on rectified results obtained by different methods. }
\vspace{1mm}
\setlength{\tabcolsep}{0.7mm}{
\begin{tabular}{ccccc}
\hline
\small Methods& Bukhar\cite{bukhari2013automatic} & AlemnFlores\cite{aleman2014automatic}& Rong\cite{rong2016radial}& Ours\\
\hline
PSNR& 9.3391& 10.23& 12.92& {\bf27.61}\\
SSIM& 0.1782& 0.2587& 0.3154& {\bf0.8746}\\
RPE&164.7 &125.4 & 121.6& {\bf0.4761}\\
\hline
\end{tabular}}
\label{table:1}
\vspace{-5mm}
\end{table}
\begin{figure}[t]
\centering
\vspace{0mm}
\begin{rotate}{90}
\hspace{1ex} D-Wireframe
\end{rotate}
\hspace{0.1ex}
\subfigure[]{
\begin{minipage}[b]{0.15\linewidth}
\includegraphics[width=1\linewidth]{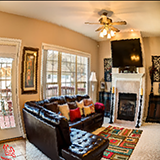}
\includegraphics[width=1\linewidth]{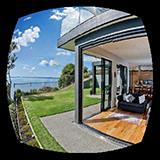}
\end{minipage}
}
\hspace{-1.9ex}
\subfigure[]{
\begin{minipage}[b]{0.15\linewidth}
\includegraphics[width=1\linewidth]{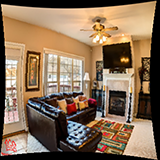}
\includegraphics[width=1\linewidth]{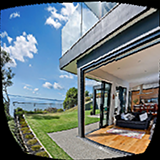}
\end{minipage}
}
\hspace{-1.9ex}
\subfigure[]{
\begin{minipage}[b]{0.15\linewidth}
\includegraphics[width=1\linewidth]{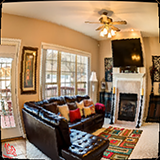}
\includegraphics[width=1\linewidth]{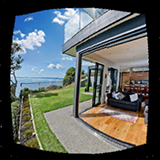}
\end{minipage}
}
\hspace{-1.9ex}
\subfigure[]{
\begin{minipage}[b]{0.15\linewidth}
\includegraphics[width=1\linewidth]{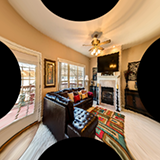}
\includegraphics[width=1\linewidth]{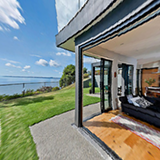}
\end{minipage}
}
\hspace{-1.9ex}
\subfigure[]{
\begin{minipage}[b]{0.15\linewidth}
\includegraphics[width=1\linewidth]{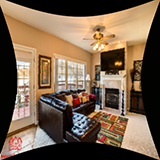}
\includegraphics[width=1\linewidth]{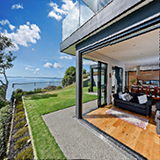}
\end{minipage}
}
\hspace{-1.9ex}
\subfigure[]{
\begin{minipage}[b]{0.15\linewidth}
\includegraphics[width=1\linewidth]{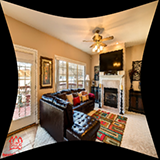}
\includegraphics[width=1\linewidth]{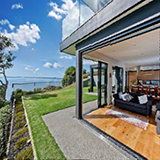}
\end{minipage}
} \\
\centering
\vspace{-4.5ex}
\begin{rotate}{90}
\hspace{1ex} Fish-SUNGCG
\end{rotate}
\hspace{0.1ex}
\subfigure[][{\scriptsize Input}]{
\begin{minipage}[b]{0.15\linewidth}
\includegraphics[width=1\linewidth]{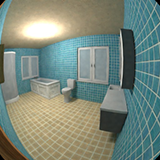}
\includegraphics[width=1\linewidth]{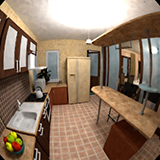}
\end{minipage}
}
\hspace{-1.9ex}
\subfigure[][{\scriptsize Bukhari}]{
\begin{minipage}[b]{0.15\linewidth}
\includegraphics[width=1\linewidth]{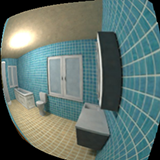}
\includegraphics[width=1\linewidth]{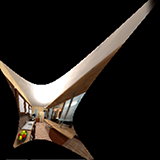}
\end{minipage}
}
\hspace{-1.9ex}
\subfigure[][{\scriptsize AlemnFlores}]{
\begin{minipage}[b]{0.15\linewidth}
\includegraphics[width=1\linewidth]{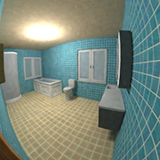}
\includegraphics[width=1\linewidth]{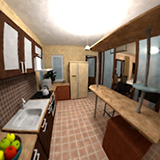}
\end{minipage}
}
\hspace{-1.9ex}
\subfigure[][{\scriptsize Rong}]{
\begin{minipage}[b]{0.15\linewidth}
\includegraphics[width=1\linewidth]{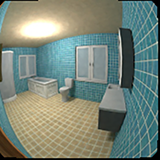}
\includegraphics[width=1\linewidth]{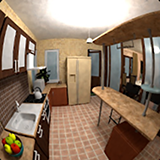}
\end{minipage}
}
\hspace{-1.9ex}
\subfigure[][{\scriptsize Ours}]{
\begin{minipage}[b]{0.15\linewidth}
\includegraphics[width=1\linewidth]{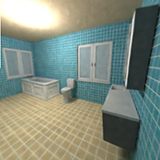}
\includegraphics[width=1\linewidth]{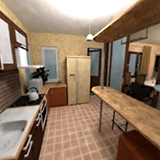}
\end{minipage}
}
\hspace{-1.9ex}
\subfigure[][{\scriptsize GT}]{
\begin{minipage}[b]{0.15\linewidth}
\includegraphics[width=1\linewidth]{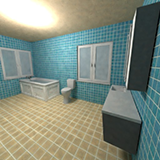}
\includegraphics[width=1\linewidth]{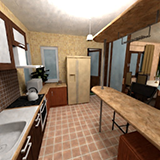}
\end{minipage}
}
\caption{Qualitative rectification comparison on D-Wireframe dataset and fish-SUNGCG dataset. From left to right, the input fisheye images, the ground truth, results of three state-of-the-art methods (Bukhari~\cite{bukhari2013automatic}, AlemnFlores~\cite{aleman2014automatic}, Rong~\cite{rong2016radial}), our results.}
\vspace{-7mm}
\label{fig:parametric}
\end{figure}

\begin{figure*}[t]
\centering
\subfigure[][{\scriptsize Input}]{
\begin{minipage}[b]{0.09\linewidth}
\includegraphics[width=1\linewidth]{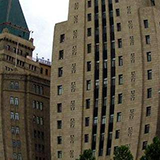}
\includegraphics[width=1\linewidth]{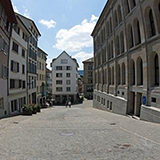}
\end{minipage}
}
\hspace{-1.9ex}
\subfigure[][{\scriptsize Bukhari~\cite{bukhari2013automatic}}]{
\begin{minipage}[b]{0.09\linewidth}
\includegraphics[width=1\linewidth]{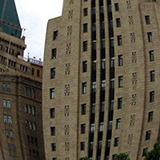}
\includegraphics[width=1\linewidth]{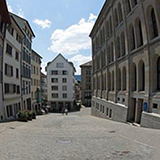}
\end{minipage}
}
\hspace{-1.9ex}
\subfigure[][{\scriptsize\centering{AlemnFlores~\cite{aleman2014automatic}}}]
{
\begin{minipage}[b]{0.09\linewidth}
\includegraphics[width=1\linewidth]{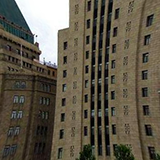}
\includegraphics[width=1\linewidth]{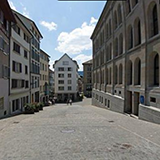}
\end{minipage}
}
\hspace{-1.9ex}
\subfigure[][{\scriptsize Rong~\cite{rong2016radial}}]{
\begin{minipage}[b]{0.09\linewidth}
\includegraphics[width=1\linewidth]{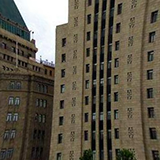}
\includegraphics[width=1\linewidth]{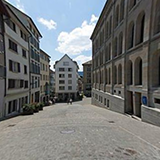}
\end{minipage}
}
\hspace{-1.9ex}
\subfigure[][{\scriptsize Ours}]{
\begin{minipage}[b]{0.09\linewidth}
\includegraphics[width=1\linewidth]{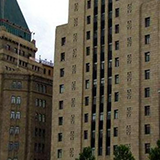}
\includegraphics[width=1\linewidth]{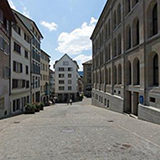}
\end{minipage}
}
\subfigure[][{\scriptsize Input}]{
\begin{minipage}[b]{0.09\linewidth}
\includegraphics[width=1\linewidth]{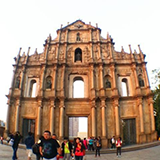}
\includegraphics[width=1\linewidth]{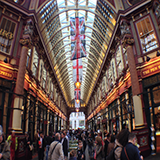}
\end{minipage}
}
\hspace{-1.9ex}
\subfigure[][{\scriptsize Bukhari~\cite{bukhari2013automatic}}]{
\begin{minipage}[b]{0.09\linewidth}
\includegraphics[width=1\linewidth]{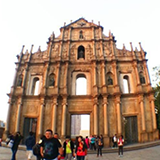}
\includegraphics[width=1\linewidth]{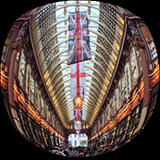}
\end{minipage}
}
\hspace{-1.9ex}
\subfigure[][{\scriptsize\centering{AlemnFlores~\cite{aleman2014automatic}}}]
{
\begin{minipage}[b]{0.09\linewidth}
\includegraphics[width=1\linewidth]{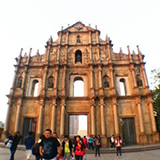}
\includegraphics[width=1\linewidth]{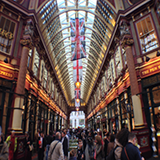}
\end{minipage}
}
\hspace{-1.9ex}
\subfigure[][{\scriptsize Rong~\cite{rong2016radial}}]{
\begin{minipage}[b]{0.09\linewidth}
\includegraphics[width=1\linewidth]{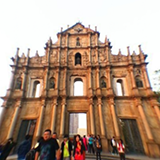}
\includegraphics[width=1\linewidth]{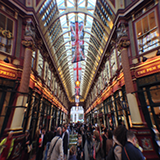}
\end{minipage}
}
\hspace{-1.9ex}
\subfigure[][{\scriptsize Ours}]{
\begin{minipage}[b]{0.09\linewidth}
\includegraphics[width=1\linewidth]{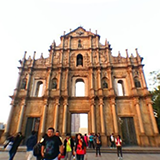}
\includegraphics[width=1\linewidth]{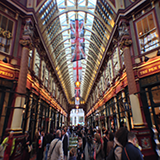}
\end{minipage}
}
\caption{Qualitative rectification comparison on fisheye images taken by real fisheye cameras.
}
\vspace{-12pt}
\label{fig:real}
\end{figure*}
\begin{figure*}[t]
\centering
\subfigure[][{Input}]{
\begin{minipage}[b]{0.12\linewidth}
\includegraphics[width=1\linewidth]{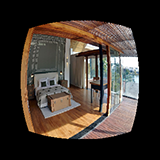}
\includegraphics[width=1\linewidth]{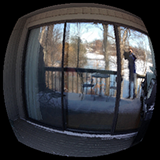}
\end{minipage}
}
\hspace{-1.9ex}
\subfigure[][\centering{w/o CVC\&MSP}]{
\begin{minipage}[b]{0.12\linewidth}
\includegraphics[width=1\linewidth]{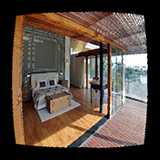}
\includegraphics[width=1\linewidth]{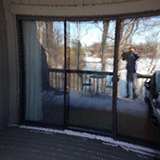}
\end{minipage}
}
\hspace{-1.9ex}
\subfigure[][\centering{w/o CVC }]{
\begin{minipage}[b]{0.12\linewidth}
\includegraphics[width=1\linewidth]{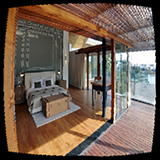}
\includegraphics[width=1\linewidth]{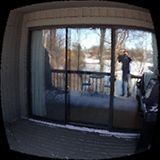}
\end{minipage}
}
\hspace{-1.9ex}
\subfigure[][\centering{ w/o MSP}]{
\begin{minipage}[b]{0.12\linewidth}
\includegraphics[width=1\linewidth]{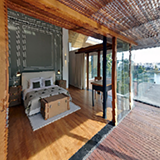}
\includegraphics[width=1\linewidth]{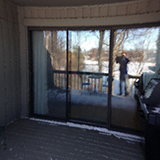}
\end{minipage}
}
\hspace{-1.9ex}
\subfigure[][{ only RGB}]{
\begin{minipage}[b]{0.12\linewidth}
\includegraphics[width=1\linewidth]{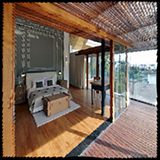}
\includegraphics[width=1\linewidth]{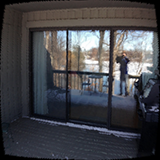}
\end{minipage}
}
\hspace{-1.9ex}
\subfigure[][{only Line}]{
\begin{minipage}[b]{0.12\linewidth}
\includegraphics[width=1\linewidth]{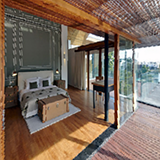}
\includegraphics[width=1\linewidth]{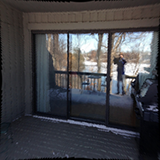}
\end{minipage}
}
\hspace{-1.9ex}
\subfigure[][{ Ours}]{
\begin{minipage}[b]{0.12\linewidth}
\includegraphics[width=1\linewidth]{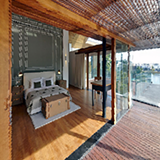}
\includegraphics[width=1\linewidth]{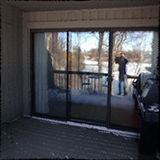}
\end{minipage}
}
\hspace{-1.9ex}
\subfigure[][{ Ground Truth}]{
\begin{minipage}[b]{0.12\linewidth}
\includegraphics[width=1\linewidth]{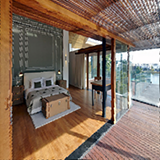}
\includegraphics[width=1\linewidth]{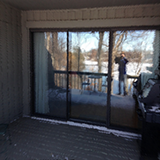}
\end{minipage}
}
\caption{Ablation experiments to verify effectiveness of the curvature constraint(CVC) and the local perception(LP) of fisheye image.}
\vspace{-10pt}
\label{fig:peeltest}
\end{figure*}

We also follow the evaluation metrics PSNR, SSIM as well as reprojection errors (RPE) for qualitative evaluation of all rectified fisheye images on our test set, as reported in Tab.~{\ref{table:1}}. From the evaluation results, it demonstrate that no matter in image rectification or in structure maintenance, our method is obviously superior to other methods and has achieved the highest score on PSNR, SSIM as well as RPE. It is worth mentioning that the reprojection error of the whole image calculated by our method is within one pixel, while other methods far behind us. Just for the reason that the estimated fisheye distortion parameters are highly consistent with the true parameters, our reprojection error can be controlled in such small range, resulting in the best rectification visual effect.

For the rectification effects in visual effect, we visualize the rectification effects by our method and start-of-the-art methods~\cite{bukhari2013automatic,aleman2014automatic,rong2016radial} on the test set of D-Wireframe and Fish-SUNCG collections respectively, as shown in Fig.~{\ref{fig:parametric}}.
For the D-Wireframe, we selected special images of different types of fisheye distortion, such as typical full-frame fisheye image, full circle fisheye image and drum fisheye image, and the results show that our method has excellent rectification effect in visual effects, while other methods can not satisfy the needs of correcting various distortion effects of fisheye images. For the Fish-SUNCG, our network also achieves the best rectification performance and the rectified fisheye image is more closer to ground truth.

Finally, we add an additional set of comparison experiments to rectify the fisheye image in real world for the generalization performance of the proposed network, as shown in Fig.~{\ref{fig:real}}. Although we do not have the internal camera parameters for taking these fisheye images, it demonstrate that our method has excellent rectification performance even for real fisheye images from the visual effect and prove that our network has higher generalization ability.

\subsection{Ablation Study}
In this section, we mainly analyze the validity of our network structures involving geometric learning, including the fourth dimensional input of concatenating the input image and detected {\em distorted lines} map in LPE, curvature constraint (CVC), as well as multi-scale perception (MSP) for locally and globally estimating the distortion parameters. As shown in Fig.~{\ref{fig:peeltest}},
once there are lack of CVC or MSP, the rectification effect of the network will become unstable, meanwhile problems of over-rectified and under-rectified will occur, and the network with RGB fisheye image and one-dimensional line map input performs best in rectification performance.


\begin{table}[t]
\centering
\caption{Ablation study to evaluate the rectified image quality of PSNR, SSIM and reprojection error (RPE).}
\vspace{1mm}
\setlength{\tabcolsep}{1mm}{
\begin{tabular}{c|c|ccc}
\hline
&Methods& PSNR& SSIM &RPE\\
\hline
\multirow{3}{40pt}{\centering {\em Loss \\ Strategy}} &
{ w/o} CVC\&MSP& 13.05& 0.4222 &78.32\\
&{ w/o} CVC& 19.17& 0.7074& 4.326\\
&{ w/o} MSP& 19.78& 0.6999& 3.175\\
\hline
\multirow{3}{*}{{\em Input}}&
only RGB& {21.35}& {0.7158} &{ 1.872}\\
&only Line& {22.41}& {0.7369} &{ 1.568}\\
&Ours& {\bf27.61}& {\bf0.8746} &{\bf 0.4761}\\
\hline
\end{tabular}}
\label{table:2}
\vspace{-5mm}
\end{table}

For qualitative evaluation, we evaluated the quality of rectified images on PSNR, SSIM and RPE, as shown in Tab.~{\ref{table:2}}. It demonstrates the ability level differences of rectification more intuitively, and proves that CVC, MSP, as well as the four-dimensional input (RGB+Line) do play critical roles in our network. According to our analysis, it is probably that the network learned high-level structural information from the {\em distorted lines} that boost the effects.
In addition, the worst results in this experiment is still better than state-of-the-art methods. It also proves the scientificity and reasonability of our network.

\vspace{-1mm}
\section{Conclusion}
\vspace{-1mm}
In this paper, we proposed a network that utilize line constraints to calibrate the fisheye lenses and eliminate the distortion effects automatically from single image. To train the network, we reuse the existing datasets that have rich $2$D and $3$D geometric information to generate the a synthetic dataset for fisheye calibration.
The proposed method takes the advantages of geometry aware deep features,
curvature constraints and multi-scale perception blocks to achieve the best performance compared to  the state-of-the-art methods, both qualitatively and quantitatively.

\noindent{\bf Acknowledgment:} This work is supported by NSFC-projects under contracts
No.61771350 and 61842102. Nan Xue is supported by China Scholarship Council. Thanks for the support and help from tutors and seniors.

{\small
\bibliographystyle{ieee}
\bibliography{egbib}

\begin{thebibliography}{10}\itemsep=-1pt

\bibitem{aleman2014automatic}
M.~Alem{\'a}n-Flores, L.~Alvarez, L.~Gomez, and D.~Santana-Cedr{\'e}s.
\newblock Automatic lens distortion correction using one-parameter division
  models.
\newblock {\em IPOL}, 4:327--343, 2014.

\bibitem{alhaija2018augmented}
H.~A. Alhaija, S.~K. Mustikovela, L.~Mescheder, A.~Geiger, and C.~Rother.
\newblock Augmented reality meets computer vision: Efficient data generation
  for urban driving scenes.
\newblock {\em IJCV}, 126(9):961--972, 2018.

\bibitem{aliaga2001accurate}
D.~G. Aliaga.
\newblock Accurate catadioptric calibration for real-time pose estimation in
  room-size environments.
\newblock In {\em ICCV}, 2001.

\bibitem{barreto2009automatic}
J.~Barreto, J.~Roquette, P.~Sturm, and F.~Fonseca.
\newblock Automatic camera calibration applied to medical endoscopy.
\newblock In {\em BMVC}, 2009.

\bibitem{barreto2005geometric}
J.~P. Barreto and H.~Araujo.
\newblock Geometric properties of central catadioptric line images and their
  application in calibration.
\newblock {\em PAMI}, 27(8):1327--1333, 2005.

\bibitem{basu1995alternative}
A.~Basu and S.~Licardie.
\newblock Alternative models for fish-eye lenses.
\newblock {\em Pattern recognition letters}, 16(4):433--441, 1995.

\bibitem{bazin2007rectangle}
J.-C. Bazin, I.~Kweon, C.~Demonceaux, and P.~Vasseur.
\newblock Rectangle extraction in catadioptric images.
\newblock 2007.

\bibitem{Bertasius2014DeepEdge}
G.~Bertasius, J.~Shi, and L.~Torresani.
\newblock Deepedge: A multi-scale bifurcated deep network for top-down contour
  detection.
\newblock In {\em CVPR}, 2014.

\bibitem{bertozzi2000vision}
M.~Bertozzi, A.~Broggi, and A.~Fascioli.
\newblock Vision-based intelligent vehicles: State of the art and perspectives.
\newblock {\em RAS}, 32(1):1--16, 2000.

\bibitem{blender2014blender}
{Blender Online Community}.
\newblock Blender - a 3d modelling and rendering package.
\newblock Blender Foundation, Blender Institute Amsterdam, 2014.

\bibitem{brauer2001new}
C.~Brauer-Burchardt and K.~Voss.
\newblock A new algorithm to correct fish-eye-and strong
  wide-angle-lens-distortion from single images.
\newblock In {\em ICIP}, 2001.

\bibitem{bukhari2013automatic}
F.~Bukhari and M.~N. Dailey.
\newblock Automatic radial distortion estimation from a single image.
\newblock {\em JMIV}, 45(1):31--45, 2013.

\bibitem{devernay2001straight}
F.~Devernay and O.~Faugeras.
\newblock Straight lines have to be straight.
\newblock {\em MVA}, 13(1):14--24, 2001.

\bibitem{Grossberg2001A}
M.~D. Grossberg and S.~K. Nayar.
\newblock A general imaging model and a method for finding its parameters.
\newblock In {\em ICCV}, 2001.

\bibitem{he2016deep}
K.~He, X.~Zhang, S.~Ren, and J.~Sun.
\newblock Deep residual learning for image recognition.
\newblock In {\em CVPR}, 2016.

\bibitem{huang20176}
J.~Huang, Z.~Chen, D.~Ceylan, and H.~Jin.
\newblock 6-dof vr videos with a single 360-camera.
\newblock In {\em VR}, 2017.

\bibitem{huang2018learning}
K.~Huang, Y.~Wang, Z.~Zhou, T.~Ding, S.~Gao, and Y.~Ma.
\newblock Learning to parse wireframes in images of man-made environments.
\newblock In {\em CVPR}, 2018.

\bibitem{Hwang2015Pixel}
J.~J. Hwang and T.~L. Liu.
\newblock Pixel-wise deep learning for contour detection.
\newblock In {\em ICLR}, 2015.

\bibitem{kannala2006generic}
J.~Kannala and S.~S. Brandt.
\newblock A generic camera model and calibration method for conventional,
  wide-angle, and fish-eye lenses.
\newblock {\em PAMI}, 28(8):1335--1340, 2006.

\bibitem{DSN2015}
C.~Lee, S.~Xie, P.~W. Gallagher, Z.~Zhang, and Z.~Tu.
\newblock Deeply-supervised nets.
\newblock In {\em AISTATSZ}, 2015.

\bibitem{liu2017richer}
Y.~Liu, M.-M. Cheng, X.~Hu, K.~Wang, and X.~Bai.
\newblock Richer convolutional features for edge detection.
\newblock In {\em CVPR}, 2017.

\bibitem{melo2013unsupervised}
R.~Melo, M.~Antunes, J.~P. Barreto, G.~Falcao, and N.~Goncalves.
\newblock Unsupervised intrinsic calibration from a single frame using a.
\newblock In {\em ICCV}, 2013.

\bibitem{miyamoto1964fish}
K.~Miyamoto.
\newblock Fish eye lens.
\newblock {\em JOSA}, 54(8):1060--1061, 1964.

\bibitem{Newell2016Stacked}
A.~Newell, K.~Yang, and J.~Deng.
\newblock Stacked hourglass networks for human pose estimation.
\newblock In {\em ECCV}, 2016.

\bibitem{Peter2006Self}
S.~R. Peter~Sturm.
\newblock Self-calibration of a general radially symmetric.
\newblock In {\em ECCV}, 2006.

\bibitem{Piotr2015Fast}
D.~Piotr and Z.~C~Lawrence.
\newblock Fast edge detection using structured forests.
\newblock {\em PAMI}, 37(8):1558--1570, 2015.

\bibitem{rong2016radial}
J.~Rong, S.~Huang, Z.~Shang, and X.~Ying.
\newblock Radial lens distortion correction using convolutional neural networks
  trained with synthesized images.
\newblock In {\em ACCV}, pages 35--49, 2016.

\bibitem{ScaramuzzaMS06}
D.~Scaramuzza, A.~Martinelli, and R.~Siegwart.
\newblock A flexible technique for accurate omnidirectional camera calibration
  and structure from motion.
\newblock In {\em ICVS}, 2006.

\bibitem{song2016ssc}
S.~Song, F.~Yu, A.~Zeng, A.~X. Chang, M.~Savva, and T.~Funkhouser.
\newblock Semantic scene completion from a single depth image.
\newblock In {\em CVPR}, 2017.

\bibitem{Sturm2004A}
P.~Sturm and S.~Ramalingam.
\newblock A generic concept for camera calibration.
\newblock In {\em ECCV}, 2004.

\bibitem{szeliski1997creating}
R.~Szeliski and H.-Y. Shum.
\newblock Creating full view panoramic image mosaics and environment maps.
\newblock In {\em SIGGRAPH}, 1997.

\bibitem{toepfer2007unifying}
C.~Toepfer and T.~Ehlgen.
\newblock A unifying omnidirectional camera model and its applications.
\newblock In {\em ICCV}, 2007.

\bibitem{xie2015holistically}
S.~Xie and Z.~Tu.
\newblock Holistically-nested edge detection.
\newblock In {\em ICCV}, 2015.

\bibitem{xiong1997creating}
Y.~Xiong and K.~Turkowski.
\newblock Creating image-based vr using a self-calibrating fisheye lens.
\newblock In {\em CVPR}, 1997.

\bibitem{yin2018fisheyerecnet}
X.~Yin, X.~Wang, J.~Yu, M.~Zhang, P.~Fua, and D.~Tao.
\newblock Fisheyerecnet: A multi-context collaborative deep network for fisheye
  image rectification.
\newblock {\em ECCV}, 2018.

\bibitem{zhang2015line}
M.~Zhang, J.~Yao, M.~Xia, K.~Li, Y.~Zhang, and Y.~Liu.
\newblock Line-based multi-label energy optimization for fisheye image
  rectification and calibration.
\newblock In {\em CVPR}, 2015.

\end{thebibliography}
}

\end{document}